\documentclass[preprint,12pt]{elsarticle}

\usepackage{graphicx}
\usepackage{amsmath,bm}
\usepackage[table,xcdraw]{xcolor}
\usepackage[toc,page]{appendix}
\usepackage{natbib}
\usepackage[utf8]{inputenc}
\usepackage{mathtools}
\usepackage{hyperref}
\biboptions{sort&compress}
\usepackage[normalem]{ulem}
\usepackage[dvipsnames]{xcolor}
\definecolor{MyGreen}{rgb}{0, 0.65, 0}
\usepackage{placeins}
\makeatletter
\AtBeginDocument{%
  \expandafter\renewcommand\expandafter\subsection\expandafter
    {\expandafter\@fb@secFB\subsection}%
  \newcommand\@fb@secFB{\FloatBarrier
    \gdef\@fb@afterHHook{\@fb@topbarrier \gdef\@fb@afterHHook{}}}%
  \g@addto@macro\@afterheading{\@fb@afterHHook}%
  \gdef\@fb@afterHHook{}%
}
\makeatother

\usepackage{graphicx,epsfig,amssymb,amsmath,amsthm,amscd,amsfonts,bm}
\usepackage{amssymb}

\journal{Computer Methods in Applied Mechanics and Engineering}

\begin{document}

\begin{frontmatter}

\title{Learning momenta for Hamiltonian-based models}

\author[inst1]{Ignacio Puiggros T.}

\author[inst1]{A. Srikantha Phani}

\affiliation[inst1]{organization={Department of Mechanical Engineering, The University of British Columbia},
            addressline={2054-6250 Applied Science Lane}, 
            city={Vancouver},
            postcode={V6T 1Z4}, 
            state={British Columbia},
            country={Canada}}

\begin{abstract}
There has been increasing interest in methodologies that incorporate physics priors into neural network architectures to enhance their modeling capabilities. Among these methodologies, Hamiltonian neural networks (HNNs) and their variants have gained considerable attention. These architectures explicitly encode Hamiltonian mechanics both in their structure and loss function, resulting in desirable properties and additional modeled information. A crucial limitation of HNNs is their reliance on data on the target system's generalized momenta. This severely hampers their applicability when the analytical form of the generalized momenta is unknown, which frequently is the case in real-world applications. The goal of the present work is to develop a general extension of Hamiltonian-based architectures that addresses this issue by simultaneously learning the system's generalized momenta, Hamiltonian structure, and dynamics from data. The proposed architecture learns the generalized momenta when their analytical form is unknown, and only (potentially non-sequential) data on the state and its derivatives is available. A dissipative mass-spring system, a double pendulum, and a rolling disk under nonholonomic constraints are considered as canonical examples to assess the performance of the proposed architecture. 

\end{abstract}


\begin{highlights}
\item System dynamics, Hamiltonian, and generalized momenta are modeled simultaneously.  
\item The architecture requires data only on the state and its first two time derivatives.
\item The extension enables Hamiltonian-based architectures when the momenta are unknown.
\item Tested systems include non-Hermitian degeneracies and nonholonomic constraints.
\end{highlights}


\end{frontmatter}


\section{Introduction}
\label{Introduction}
Data-driven approaches for the modeling of dynamical systems have seen a rise in popularity, with many works implementing methodologies such as sparse identification of nonlinear dynamics (SINDy) \cite{Sindy1,Sindy2,Sindy3}, dynamic mode decomposition (DMD) \cite{DMD1,DMD2,DMD3}, symbolic modeling \cite{Symbolic}, and neural networks \cite{PINN,HNN,GHNN,Nonholonomic-Robot}.
The latter have been particularly successful, as they provide a versatile framework for modeling directly from
data that is well equipped to deal with high-dimensional problems \cite{weinan2021dawning}.
An increasingly common approach is to incorporate physics priors in data-driven strategies \cite{Sindy3,DMD3,PINN}, improving modeling capability, computation time, and reducing model complexity. The incorporation of physics priors can be done explicitly by embedding them in the architecture itself, in the loss function, or both. \\


A family of methodologies with physics priors that has seen a growth in popularity are Hamiltonian-based models, and in particular, Hamiltonian neural networks (HNN) \cite{HNN}. HNNs explicitly use Hamiltonian mechanics as a prior in their formulation to enhance their modeling power. Greydanus et al. \cite{HNN} proposed to model Hamiltonian systems by learning the Hamiltonian itself as the single output of a neural network, effectively encoding the dynamics of the system. Once the Hamiltonian is learned, autodifferentiation can be used to obtain its gradient, which together with Hamilton’s equations gives the dynamics of the system. This architecture does not only simplify the function to be learned by the network when compared to the full coordinate-derivative map, but also guarantees that the learned Hamiltonian is conserved. HNNs were found to outperform standard neural network modeling approaches, achieving higher precision and more accurately conserving energy.\\

Hamiltonian neural networks have been greatly expanded upon, extending their applicability to a wider class of problems. Desai et al. \cite{Port-HNN} propose port-Hamiltonian neural networks (pHNNs), including both dissipation and external forces in the formulation. 
Another improvement is developed by Han et al. \cite{AHNN}, introducing Adaptable HNNs (AHNN), with more inputs in the first layer to account for variations in the system parameters. More recently, we proposed \cite{PuiggrosNHHNN} an architecture that extends Hamiltonian neural networks to systems under nonholonomic constraints, simultaneously learning the Hamiltonian, constraints, and Lagrange multipliers of the system.\\

Although HNNs and their extensions have shown promising results, they fundamentally rely on the availability of data on the generalized momenta and their time derivatives. This is highly prohibitive in real-world applications, where the generalized momenta are often unknown nonlinear functions of the velocities, severely limiting the applicability of these models. Several existing approaches offer alternative ways to address this limitation, including Lagrangian Neural Networks (LNNs) \cite{LNN}, Neural Symplectic Forms (NSFs) \cite{NSF}, Hamiltonian Generative Networks (HGNs) \cite{HGN}, and autoencoder-based architectures like the one proposed by Bertalan et al. \cite{bertalan}. While these approaches show notable results, they continue to face some key challenges. \\

Lagrangian Neural Networks \cite{LNN} represent an alternative approach capable of modeling the system when only data on the state and its derivatives is available by learning the Lagrangian of the system. However, this formulation cannot fully replace Hamiltonian-based approaches, as it does not immediately provide a model for the Hamiltonian of the system, numerical integration is harder due to the equations being of second order, the learned dynamics are not immediately symplectic, and the computation of the Hessian of the Lagrangian significantly increases computational cost. \\

Chen et al. \cite{NSF} proposed Neural Symplectic Forms, where the coordinate-free Hamiltonian formulation is used to model Hamiltonian systems using data on the state and its derivatives. They learn both the Hamiltonian of the system and the corresponding symplectic 2-form as a neural network, enabling their methodology when only data on the state and its time derivatives are available. Although the authors show promising results and their architecture conserves energy by construction, the proposed approach is not without limitations. First, their methodology does not provide a representation of the generalized momenta of the system, losing information on the generalized forces, impeding the use of canonical transformations, and making it no longer readily compatible with traditional symplectic numerical integrators. Additionally, NSFs come with an increased computational cost, as they require the computation of the exterior derivative and the inverse of the symplectic matrix, both of which scale poorly with the dimensions of the target system. Lastly, their architecture is not readily compatible with available Hamiltonian-based approaches that require data on phase space.\\

In a similar vein, Toth et al. \cite{HGN} propose the use of Hamiltonian Generative Networks, learning the phase space coordinates of a system from general observations as the latent space of a variational autoencoder \cite{VAE}, while simultaneously learning the Hamiltonian of the system in the encoded space. The authors show promising results for systems with simple generalized momenta, such as a pendulum or the three-body problem. Although useful for dealing with high-dimensional representations of a system with unknown phase space coordinates, their architecture has several limitations. Their model assumes data consisting of regularly sampled trajectories, hampering its applicability for irregularly sampled or non-sequential datasets. Additionally, the inherent stochastic nature of their autoencoder, together with the computation of gradients over many consecutive passes of the network, may lead to convergence issues and loss of precision when modeling highly complex and potentially chaotic systems. Furthermore, their approach is not tested on more complex systems, such as systems with nonholonomic constraints or nontrivial generalized momenta.\\

Similarly, Bertalan et al. \cite{bertalan} proposed to learn the phase space of a system from general observations as the latent space of an autoencoder. A separate network is simultaneously trained to learn the Hamiltonian as a function of the encoded variables, effectively modeling the dynamics of the system. Although the authors show promising results, they only test their architecture on systems with trivial generalized momenta (i.e. velocity times a constant), and do not assess performance with more complex systems such as systems with dissipation, nonholonomic constraints, or non-Hermitian degeneracies. Furthermore, their approach requires additional terms in the loss function to ensure stability. These additional terms potentially increase optimization complexity, extend training times, and make implementation on available Hamiltonian-based architectures more challenging.\\

The main contribution of this work is a general extension to Hamiltonian-based architectures that simultaneously learns the target system's Hamiltonian, generalized momenta, dynamics, and mathematical structure over a region of interest in state space. We propose a single-network extension that explicitly learns the generalized momenta as a function of generalized position and velocity, whose output is then used in the Hamiltonian formulation. Our methodology assumes only (potentially non-sequential) data on the state and its derivatives are available. The resulting framework is readily compatible with Hamiltonian-based architectures such as pHNNs and HNNs for systems under nonholonomic constraints, and does not require any modifications to the loss function. We evaluate the performance of the proposed extension using three canonical examples: a simple dissipative mass–spring system, a double pendulum, and a rolling disk subject to nonholonomic constraints. We note that, although systems with dissipation or nonholonomic constraints can be described with extensions of the Hamiltonian framework, they are not Hamiltonian in the traditional sense.

\section{Methodology}

\subsection{Hamiltonian mechanics}

Besides Newton's equations, there are other equivalent formulations of classical mechanics, which notably include Lagrangian and Hamiltonian mechanics. While Newtonian mechanics revolves around forces and their effect on systems, the other two define a scalar function of the parameters and state of the system, from which the equations of motion are derived through a variational approach. These functions are referred to as the Lagrangian, which typically corresponds to the kinetic energy of the system minus its potential energy, and the Hamiltonian, which is obtained through the Legendre transformation of the Lagrangian. The Hamiltonian is usually the total energy of the system, and is expressed as a function of some set of generalized coordinates $\textbf{q}$ and the corresponding generalized momenta $\textbf{p}$ \cite{goldstein}. The Hamiltonian encodes all the information of the dynamics of conservative systems, which can be derived using Hamilton’s equations:

\begin{equation} \label{Heqs}
\dot{q_{i}}=\frac{\partial H}{\partial p_{i}}, \quad \dot{p_{i}}=-\frac{\partial H}{\partial q_{i}},
\end{equation}

\noindent where $H$ is the Hamiltonian, $q_{i}$ and $p_{i}$ are the components of the vectors $\textbf{q}$ and $\textbf{p}$, respectively, and $\dot{q_{i}}$ and $\dot{p_{i}}$ symbolize total derivatives with respect to time. The vectors $\dot{\textbf{q}}$ and $\dot{\textbf{p}}$ are usually referred to as the phase velocity of the system.\\

\subsection{Momentum learning Hamiltonian neural networks}

\begin{figure}[htbp]
    \centering
    \includegraphics[width=3.54331in]{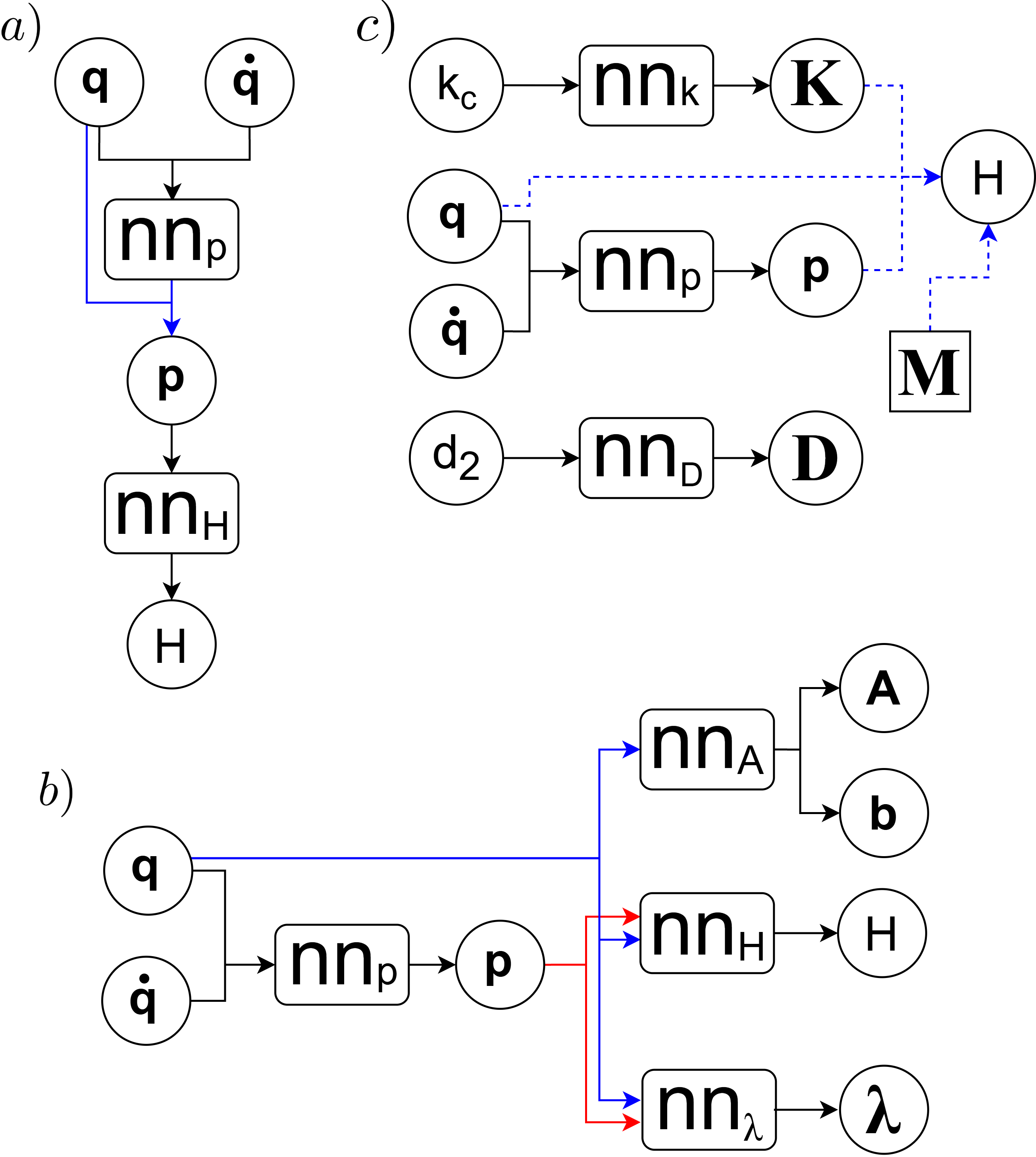}
    \caption{Network architectures incorporating the proposed extension. Solid arrows represent explicit network inputs and outputs. \textbf{a)} Two-network architecture to learn the Hamiltonian and momenta of conservative, unconstrained systems. \textbf{b)} Hamiltonian NN architecture for modeling systems under nonholonomic constraints extended to learn generalized momenta from data. \textbf{c)}  Architecture used to model the parameter-dependent dissipative mass-spring system. Dashed lines indicate quantities used to construct the Hamiltonian. $k_{c}$ and $d_{2}$ are parameters of the system.}
    \label{MHNN}
\end{figure}

We propose to modify standard Hamiltonian neural network architectures by adding a preprocessing multilayer perceptron (MLP) that learns $\textbf{p}$ as a function of $\textbf{q}$ and $\dot{\textbf{q}}$. The simplest form of the proposed extension is shown in panel \textbf{a)} of Figure \ref{MHNN}. The networks learning the generalized momenta and the Hamiltonian are trained \textit{simultaneously} without requiring any additional terms to the loss function. If implemented in the standard Hamiltonian NN formulation \cite{HNN}, the loss function is:

\begin{equation} \label{MHNNLoss}
    \mathcal{L}= \left\Vert \dot{\textbf{q}}-\frac{\partial \hat{H}}{\partial \hat{\textbf{p}}} \right\Vert^{2}_{2} + \left\Vert\hat{\dot{\textbf{p}}}+\frac{\partial \hat{H}}{\partial \textbf{q}}\right\Vert^{2}_{2}
\end{equation}\\

\noindent where the circumflex `` $\hat{\cdot}$ '' represents a modeled quantity, and  $\hat{\dot{\textbf{p}}}$ is computed as:

\begin{equation}
    \hat{\dot{\textbf{p}}}=\frac{\partial \hat{\textbf{p}}}{\partial \textbf{q}}\dot{\textbf{q}}+\frac{\partial \hat{\textbf{p}}}{\partial \dot{\textbf{q}}}\ddot{\textbf{q}}.
\end{equation} 

The set of functions $\{\hat{H},\hat{\textbf{p}}\}$ that minimizes Eq. \ref{MHNNLoss} is not unique, even if the choice of generalized coordinates $\textbf{q}$ is fixed. As a consequence, separate training runs on the same dataset can yield two valid, yet distinct, models for the Hamiltonian and momenta. This nonuniqueness of the Hamiltonian and generalized momenta is well known, and it is described in detail in the theory of canonical transformations \cite{goldstein}.\\

The proposed approach is readily compatible with currently available Hamiltonian-based architectures that model a wider class of systems, such as pHNNs  \cite{Port-HNN} or our previous extension to systems under nonholonomic constraints \cite{PuiggrosNHHNN}, requiring no further modification to the loss function. Panel \textbf{b)} in Figure \ref{MHNN} shows the full architecture when extending the methodology from \cite{PuiggrosNHHNN}. \\

The proposed approach easily allows the implementation of additional priors if available, such as learned momenta that are linear in velocities, learned momenta that are statistically independent, or a Hamiltonian that has a known form. Linear momenta and Hamiltonian priors can be directly imposed by parameterizing them accordingly, while the statistically independent momenta can be imposed through a regularization term in the loss function that penalizes high total correlation using the Kullback-Leibler (KL) divergence \cite{KLD}. The latter is not explored in the present work.\\

The implementation shown in this paper is done entirely in PyTorch, and all networks are trained using the Adam optimizer \cite{adam}. Although a different optimizer may reduce training times and improve precision of our models, we do not explore this possibility in our work. A review of other optimization methods for machine learning can be found in \cite{optreview}. The dataset is divided into a training set and a validation set to monitor overfitting. The training is done in batches, which are constructed by randomly splitting the training set before each epoch. As the training of the networks progresses, the parameters corresponding to the lowest validation loss function are stored and later used to model the target system.


\section{Three dynamical systems}
\label{TDS}

\begin{figure}[htbp]
    \centering
    \includegraphics[scale=0.5]{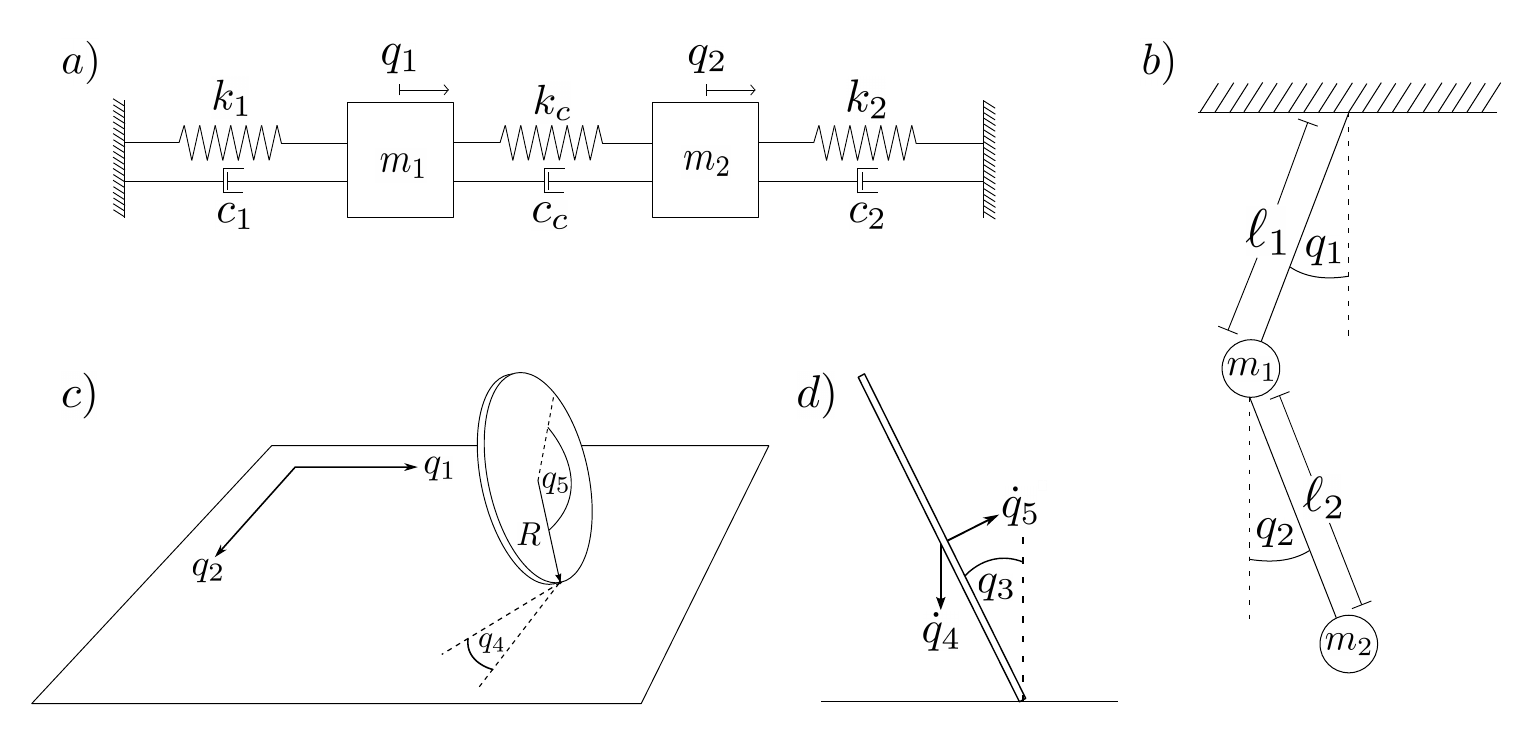}
    \caption{\textbf{a)} Notations and coordinates of the chosen dissipative mass-spring system. \textbf{b)} Choice of coordinates for the double pendulum. \textbf{c)} Disk rolling without slipping. The coordinates $q_{1}$ and $q_{2}$ indicate the horizontal position of the center of the disk, $q_{4}$ is the angle of rotation with respect to the vertical axis, and $q_{5}$ represents the rolling angle of the disk about its main symmetry axis. $R$ is the radius of the disk. \textbf{d)} $q_{3}$ is the leaning angle of the disk from its fully vertical position.}
    \label{systems}
\end{figure}

We choose three different dynamical systems, shown in Figure \ref{systems}, to assess the performance of the proposed approach: a dissipative mass-spring system, a double pendulum, and a rolling disk under nonholonomic constraints. This choice of systems allows us to evaluate our framework's compatibility with a range of Hamiltonian-based architectures, as well as its performance when dealing with nonholonomic constraints and non-Hermitian degeneracies—an aspect unexplored in similar works. We note that although all the architectures presented here assume time-independent systems, a similar approach could be used to model time-dependent systems by including time as an input to the models. \\

\begin{figure}[htbp]
    \centering
    \includegraphics[scale=0.7]{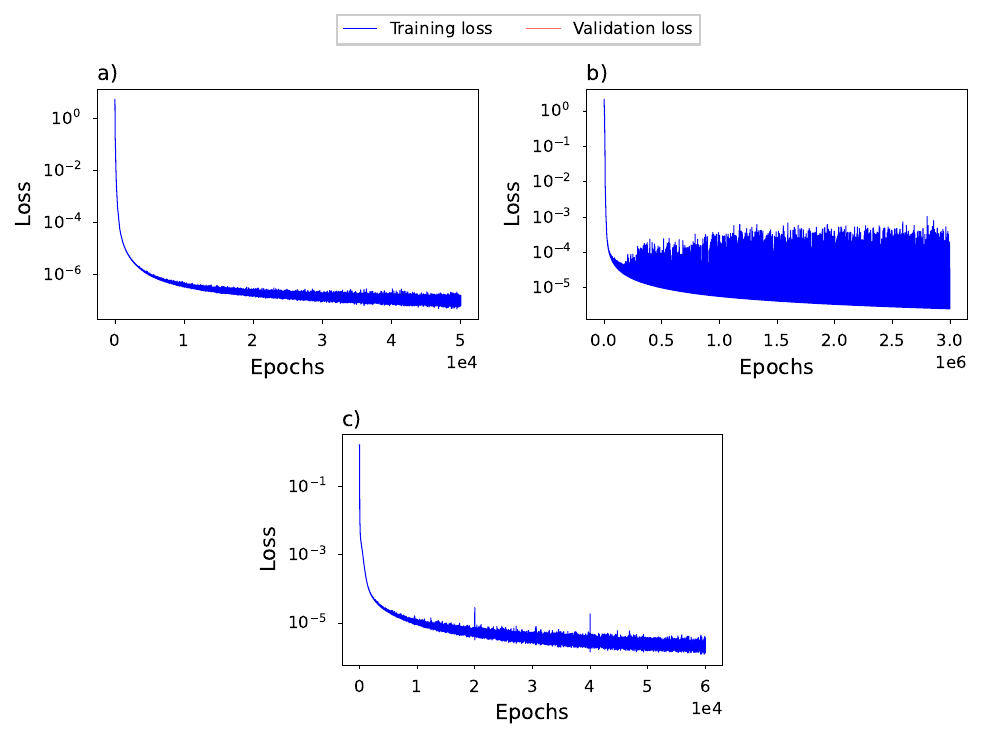}
    \caption{Evolution of the training loss function for \textbf{a)} the mass spring system, \textbf{b)} the double pendulum, and \textbf{c)} the rolling disk.}
    \label{losses}
\end{figure}

All datasets were synthetically generated, sampling uniformly distributed points from a region of interest in configuration-velocity space, and then computing accelerations analytically. The training data contains $\textbf{q}$, $\dot{\textbf{q}}$, and $\ddot{\textbf{q}}$ only. Training with nonuniformly distributed datasets is also possible as long as they are dense enough in the region of interest in state space. We do not explore this possibility and its consequences in the present work. Figure \ref{losses} shows the evolution of the loss function for all experiments.

\subsection{Dissipative mass-spring system}

We evaluate the proposed architecture with the mass-spring system shown in figure \ref{systems}. This choice of system allows us to assess the performance of the proposed extension when implemented on a dissipative variation of Greydanus' simple architecture \cite{HNN}, and when using a quadratic prior on the Hamiltonian. In order to account for dissipation, we take an approach similar to the one used for pHNNs \cite{Port-HNN}, modifying Eq. \ref{Heqs} as:

\begin{equation} \label{D}
\dot{\textbf{p}}=-\frac{\partial H}{\partial \textbf{q}} + \textbf{D}\textbf{p}.
\end{equation}

\noindent where $\textbf{D}$ is a symmetric matrix. Additionally, we choose a quadratic parameterization of the Hamiltonian as

\begin{equation} \label{Hquad}
H=\frac{1}{2} \textbf{p}^T\textbf{M}^{-1}\textbf{p} + \frac{1}{2} \textbf{q}^T\textbf{K}\textbf{q},
\end{equation}

\noindent where $\textbf{M}$ and $\textbf{K}$ are also symmetric matrices to be learned, and correspond to the mass and stiffness matrices of the system, respectively. \\


The term $\textbf{Dp}$ leads to forces in coordinate space of the form $-\textbf{C}\dot{\textbf{q}}$, where $\textbf{C}=-\textbf{D}\textbf{M}$. Because the product of two symmetric matrices is not guaranteed to be symmetric, $\textbf{C}$ might not be symmetric even if $\textbf{D}$ is. Therefore, imposing symmetry on $\textbf{D}$ does not prevent $\textbf{C}$ from having a skew-symmetric component, allowing $\textbf{Dp}$ to account for both dissipative and gyroscopic forces.\\

The chosen system also constitutes the simplest system that can present a dissipation driven degeneracy, as described by Phani in \cite{APL}. For the right system parameters, the system has a non-Hermitian degeneracy (also known as an exceptional point), i.e. a point in parameter space where both eigenvalues and eigenvectors coincide. The presence of this kind of degeneracy leads to a Berry phase in the eigenvectors of the system when circling the degeneracy in parameter space, as defined by Vanderbilt \cite{vanderbilt}: 

\begin{equation}
\phi = -\operatorname{Im}\ln\left[
\langle u_0 | u_1 \rangle
\langle u_1 | u_2 \rangle
\cdots
\langle u_{N-1} | u_0 \rangle
\right],
\end{equation}

\noindent where $\phi$ is the Berry phase gained after circling once, $N$ is the number of discrete points in parameter space used to encircle the degeneracy, and $u_i$ with $i \in \{0,1,\ldots,N-1\}$ are the corresponding eigenvectors to these $N$ points. This nonholonomic phenomenon will be used to assess the extent to which our proposed architecture is able to reproduce an emergent behavior of the system that it was not explicitly trained for. In order to achieve this, we chose the parameters shown in Table \ref{msvalues} as the degeneracy to be encircled by varying $k_c$ and $c_2$.\\

\begin{table}[htbp]
\centering
\caption{Parameters associated with the chosen non-Hermitian degeneracy of the mass-spring system.}
\label{msvalues}
\begin{tabular}{cccccccc}
\hline
\multicolumn{1}{c}{$m_{1}$} & \multicolumn{1}{c}{$m_{2}$} & \multicolumn{1}{c}{$k_{1}$} & \multicolumn{1}{c}{$k_{2}$} & \multicolumn{1}{c}{$k_{c}$} & \multicolumn{1}{c}{$c_{1}$} & \multicolumn{1}{c}{$c_{2}$} & \multicolumn{1}{c}{$c_{c}$} \\ \hline
$2$ & $1$  & $1$ & $1$ & $2.41421$ & $0$ & $2.61313$ & $0$\\ \hline
\end{tabular}
\end{table}

To allow our model to account for different system parameters, we follow a strategy similar to the one in \cite{AHNN}, by learning $\textbf{M}$ as a constant matrix, but modeling $\textbf{K}$ and $\textbf{D}$ as the outputs of two simple MLPs that take $k_c$ and $c_2$ as input, respectively. The trained models are then used to compute the eigenvectors of the system along each point in the path in parameter space encircling the degeneracy. The architecture is shown in panel \textbf{c)} of Figure \ref{MHNN}. \\

The loss function used is similar to Eq. \ref{MHNNLoss}, but is adjusted for the introduction of the term $\mathbf{Dp}$ in Eq. \ref{D}. Note that the change from Eq. \ref{MHNNLoss} to Eq. \ref{DMHNNLoss} is due to the introduction of dissipation, and not due to the preprocessing network learning the momenta:

\begin{equation} \label{DMHNNLoss}
    \mathcal{L}= \left\Vert \dot{\textbf{q}}-\frac{\partial \hat{H}}{\partial \hat{\textbf{p}}} \right\Vert^{2}_{2} + \left\Vert\hat{\dot{\textbf{p}}}+\frac{\partial \hat{H}}{\partial \textbf{q}} - \textbf{Dp}\right\Vert^{2}_{2}
\end{equation}\\

The training and architecture parameters are shown in Table \ref{mshp} of Appendix A. The dataset is created by sampling each component of state $[q_{1},q_{2},\dot{q}_{1},\dot{q}_{2}]$ uniformly in the interval $[-1.5,1.5]$, and then computing $[\ddot{q}_{1},\ddot{q}_{2}]$ analytically. For the purposes of encircling the degeneracy, $k_{c}$ and $c_{2}$ are also varied in the dataset, uniformly sampling from a unit length square in parameter space centered around the values shown in Table \ref{msvalues}. The remaining system parameters are kept fixed throughout the dataset. The total training time for the mass-spring system was 16 hours.\\

\begin{figure}[htbp]
    \centering
    \includegraphics[scale=1]{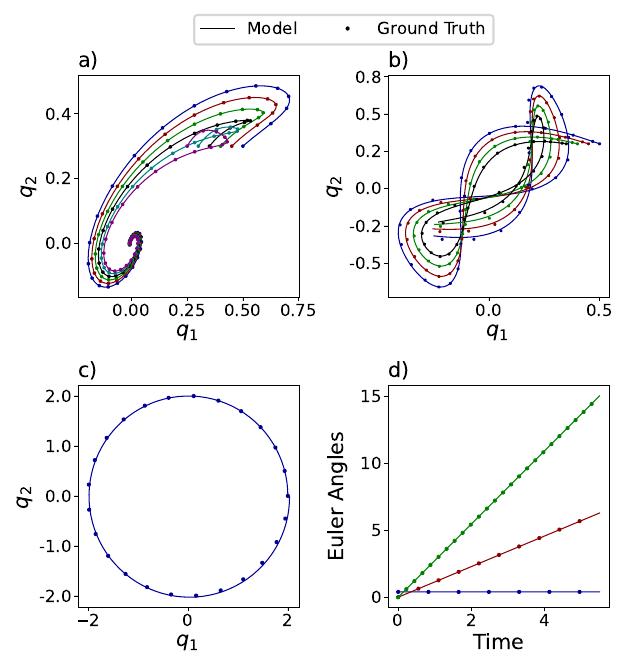}
    \caption{Modeled and ground truth trajectories for the test systems. A family of test trajectories in coordinate space of \textbf{a)} the mass-spring system and \textbf{b)} the double pendulum. \textbf{c)} Trajectory in space of the center of the disk and \textbf{d)} orientation angles of the rolling disk.  }
    \label{trajectories}
\end{figure}

The top-left panel of Figure \ref{trajectories} shows a highly accurate model reconstruction of a family of trajectories for the parameters in Table \ref{msvalues}, corresponding to initial conditions $\mathbf{z}_{0}^{(k)}=[q_1,q_2,p_1,p_2]=[\alpha_k,0.3,\alpha_k,0.3]$ where $\alpha_k \in \{0.25,0.3,0.35,0.4,0.45,0.5\}$.\\

\begin{figure}[htbp]
    \centering
    \includegraphics[scale=1]{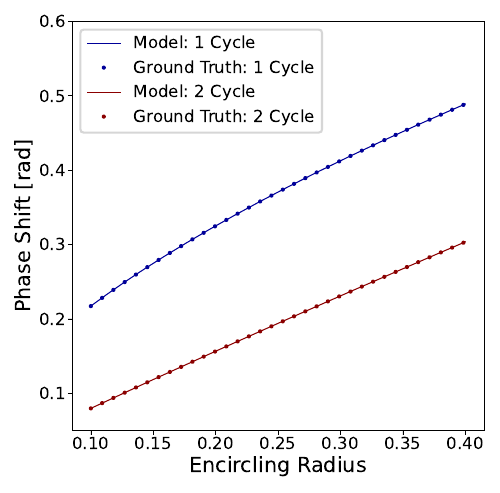}
    \caption{Phase shift as a function of the circling radius in parameter space, for one and two cycles.}
    \label{encircling}
\end{figure}

We now assess how well the proposed architecture models eigenvector phase shift when no data on the generalized momenta is available. $k_{c}$ and $d_{2}$ are slowly varied in circular paths in parameter space centered around the degeneracy. Figure \ref{encircling} shows the ground truth and modeled phase shift as a function of the circling radius in parameter space, for one and two cycles. The circling radius is defined as:

\begin{equation}
    r_{crc}=\sqrt{k_{c,crc}^{2}+d_{2,crc}^{2}}
\end{equation}

\noindent where $k_{c,crc}$ and $d_{2,crc}$ are the values of $k_{c}$ and $d_{2}$ along the circular path. The architecture recovers the phase shift behavior with high accuracy for a range of encircling paths, both for one and two cycles. We emphasize that this accurate phase shift reconstruction is not explicitly trained for, emerging as a consequence of the precise recovery of the system's dynamics.   

\subsection{Double Pendulum}

We now assess the framework's performance on the double pendulum, shown in Figure \ref{systems}. The system is chosen due to its highly nonlinear behavior and nontrivial generalized momenta, which make it an ideal test case for the proposed architecture. The analytical Hamiltonian of the system is:

\begin{equation} \label{Hquad}
\begin{gathered}
H=\frac{1}{m\ell^2(C^{2}_{\Delta}-2)^2}\left( p_1 p_2 C^{3}_{\Delta} - (\frac{p_1^{2}}{2} + p_2^{2}) C^{2}_{\Delta} - 2 p_1 p_2 C_{\Delta} + 2p_2^{2} + p_1^{2} \right) \\
-mg\ell\left(2\cos(q_{1})+\cos(q_2)\right)
\end{gathered},
\end{equation}

\noindent where $C_{\Delta}=\cos(q_2-q_1)$ and the generalized momenta are:

\begin{equation}
\label{Hquad}
\begin{aligned}
p_1 &= m\ell^2(2\dot{q}_{1} + \dot{q}_{2}C_{\Delta}),\\
p_2 &= m\ell^2(\dot{q}_{1}C_{\Delta} + \dot{q}_{2}).
\end{aligned}
\end{equation}

System parameters are chosen as [$m$, $\ell$, $g$] = [$1, 1, 9.8$]. As the system is now conservative, we use the architecture in panel \textbf{a)} of Figure \ref{MHNN} and the loss function in Eq. \ref{MHNNLoss} to simultaneously train the networks. The training and architecture parameters are shown in Table \ref{dphp} of Appendix A. The dataset is created by sampling $\textbf{q}$ uniformly in the interval $[-0.85,0.85]$, $\dot{\textbf{q}}$ in the interval $[-2.7, 2.7]$, and then computing $\ddot{\textbf{q}}$ analytically. Total training time for the double pendulum system was 19 hours. \\

The top-right panel of Figure \ref{trajectories} shows a highly accurate model reconstruction of trajectories corresponding to initial conditions  $\mathbf{z}_{0}^{(k)}=[q_{1},q_{2},\dot{q}_{1}, \dot{q}_{2}]=[\theta_k,0.3,0,0]$ where $\theta_k \in \{0.35,0.4,0.45,0.5\}$. This shows that the proposed architecture is capable of producing accurate models for the Hamiltonian formulation of highly nonlinear systems with nontrivial generalized momenta, using only data on the generalized coordinates and their derivatives.

\subsection{Rolling Disk}

As a final test case, we consider the rolling disk from \cite{PuiggrosNHHNN}, as it allows us to assess the performance of the proposed framework on nonholonomic systems and when implemented in more complex Hamiltonian-based architectures. The system consists of a disk rolling without slipping on a flat surface, a condition which needs to be expressed as a pair of nonholonomic constraints. These constraints are linear in velocities but nonlinear in the generalized coordinates, which breaks the standard Hamiltonian formulation and requires extensions to the base architecture to model the system. The thickness of the disk is assumed to be negligible. Panels \textbf{c)} and \textbf{d)} of Figure \ref{systems} show the coordinates used to describe the motion of the disk; $q_1$ and $q_2$ describe the position of the center of the disk, $q_3$ describes the leaning angle of the disk, $q_4$ describes its angle of rotation about the vertical axis in the fixed frame, and $q_5$ describes the angle of rotation about its main axis of symmetry. \\

The Hamiltonian of the system is 

\begin{equation} \label{H2}
\begin{split}
    H=&\frac{p_{1}^{2}+p_{2}^{2}}{2m}+\frac{p_{5}^{2}}{2I_{xx}}+
    \frac{(p_{4}+p_{5}\sin(q_{3}))^{2}}{\cos^{2}(q_{3})I_{xx}}
    \\&+\frac{p_{3}^{2}}{2(mR^{2}\sin^{2}(q_{3})+\frac{I_{xx}}{2})}
    +mgR\cos(q_{3}),
\end{split}
\end{equation}
%


\noindent where $I_{xx}$ is the mass moment of inertia of the disk with respect to its main axis, $m$ and $R$ are mass and radius of the disk, respectively, and $g$ is the acceleration due to gravity. System parameters are chosen as $[m,R,g]=[1,1,9.8]$. The generalized momenta of the system take the form

\begin{equation} \label{p2}
\begin{split}
&p_{1}=m\dot{q}_1, \quad
p_{2}=m\dot{q}_2, \\
&p_{3}=\dot{q}_{3}(mR^{2}\sin^{2}(q_{3})+\frac{I_{xx}}{2}),\\
& p_{4}=I_{xx}\dot{q}_{5}-I_{xx}\sin(q_{3})\dot{q}_{4}, \\
&p_{5}=I_{xx}\dot{q}_{4}\left(1-\frac{\cos^{2}(q_{3})}{2}\right)
-I_{xx}\dot{q}_{5}\sin{q_{3}}.
\end{split}
\end{equation}

Two nonholonomic constraints are needed to impose the no-slip condition to the system, and they take the form

\begin{equation} \label{c2}
\begin{split}
\dot{q}_{1}=&-R\sin(q_{4})(\dot{q}_{5}-\dot{q}_{4}\sin(q_{3}))-R\dot{q}_{3}\cos(q_{4})\cos(q_{3}),\\
\dot{q}_{2}=&R\cos(q_{4})(\dot{q}_{5}-\dot{q_{4}}\sin(q_{3}))-R\dot{q_{3}}\sin(q_{4})\cos(q_{3}).
\end{split}
\end{equation}

The base architecture and loss function are identical to the ones from \cite{PuiggrosNHHNN}, but extended as shown in panel \textbf{c)} of Figure \ref{MHNN}. The training parameters and the uniform sampling intervals for the training set are presented in Tables \ref{diskhp} and \ref{diskranges} of Appendix A, respectively. Total training time for the rolling disk experiment was 107 hours.\\

The bottom panels of Figure \ref{trajectories} shows the model reconstruction of the ground truth trajectory corresponding to the initial condition $[q_{1},q_{2},q_{3},q_{4},q_{5}, \linebreak \dot{q}_{1},\dot{q}_{2},\dot{q}_{3},\dot{q}_{4},\dot{q}_{5}]=[2, 0, 0.393, 0, 0, -0.0, 2.291, 0.0, 1.146, 2.730]$. These values are chosen as they correspond to a case of particular interest in which the center of the disk moves in a stable circle, illustrating the path-dependency that is characteristic of nonholonomic systems, as after one turn all variables return to their initial values except $\phi$, which gains an additional angle of $2.385$ rad. The test trajectory is accurately reconstructed by the model, showing that the proposed extension effectively enables complex Hamiltonian architectures for systems under nonholonomic constraints when data on the generalized momenta are unavailable.

\subsection{Learned Momenta}

We now analyze the learned generalized momenta for all three experiments. If only data on $\textbf{q}$ and its derivatives are available, then there are infinitely many pairs $[H,\textbf{p}]$ that can represent the correct dynamics of the system. This raises the question whether the learned momenta will share any similarities with the classical choice of analytical momenta for the test systems. For ease of visualization, all components of the momenta, with the exception of the angular components of the rolling disk, were centered around 0 and scaled as

\begin{equation} \label{scaling}
\begin{split}
    \textbf{p}_s=\frac{\textbf{p}-\overline{\textbf{p}}}{\overline{\vert\textbf{p}-\overline{\textbf{p}}\vert}+\epsilon},
\end{split}
\end{equation}

\noindent where $\textbf{p}_s$ is the centered and scaled momenta, the overline $\overline{\cdot}$ signals average along the signal, and $\epsilon=1e-4$ is a small positive number to avoid null division. Angular components of the rolling disk are plotted unmodified when constant along the trajectory for ease of visualization.\\

\begin{figure}[htbp]
    \centering
    \includegraphics[scale=0.67]{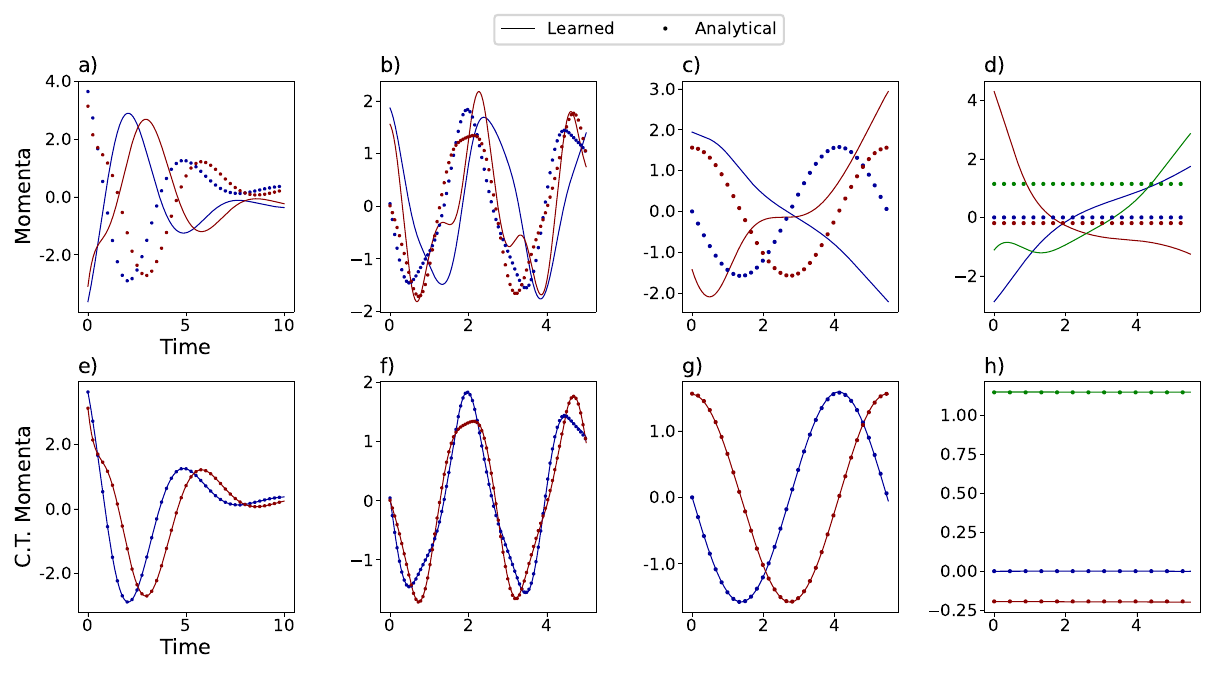}
    \caption{Comparison between learned and analytical generalized momenta for all systems. Different colors correspond to different components of the momenta, and every component, with the exception of the angular components of the rolling disk, was centered around 0 and scaled to unit mean absolute value for ease of visualization. First row shows momenta along the test trajectories for \textbf{a)} the mass-spring system, \textbf{b)} the double pendulum, \textbf{c)} spatial components of the rolling disk $[p_{1},p_{2}]$, and \textbf{d)} angular components of the rolling disk $[p_{3},p_{4},p_{5}]$. Angular components of the rolling disk are plotted unmodified when constant along the trajectory for ease of visualization. The second row compares the analytical momenta along the test trajectory with the learned momenta after being transformed as in Eq. \ref{ct}. Panel \textbf{e)} corresponds to the mass-spring system, \textbf{f)} to the double pendulum, and \textbf{g)} and \textbf{h)} to the spatial and angular components of the rolling disk momenta, respectively.
    }
    \label{momenta}
\end{figure}

The first row in Figure \ref{momenta} compares the analytical and learned momenta along the test trajectories. Panel \textbf{a)} shows that the learned and analytical momenta of the mass-spring system are approximately related by a sign change. This is consistent with the expected behavior, as the chosen parameterization together with the first part of Eq. \ref{Heqs} makes it so the learned momenta converge to $\hat{\textbf{p}}=\textbf{M}\dot{\textbf{q}}$. This means that the learned momenta and the analytical momenta $\textbf{p}_{analytical}=[2m\dot{q}_1, m\dot{q}_{2}]$ can be different up to a linear combination that depends on the learned matrix $\textbf{M}$. Panel \textbf{b)} shows that the learned and analytical momenta of the double pendulum follow the same general trends while still being distinct quantities. In panels \textbf{c)} and \textbf{d)} the learned and analytical momenta of the rolling disk no longer follow similar trends, especially the angular components, where the analytical momenta are constant along the test trajectory while none of the learned momenta are. This difference arises from the theory’s freedom of choice of Hamiltonian and momenta, allowing the model to learn quantities with a wide range of  behaviors. Although we do not explore this possibility here, this flexibility could be exploited by specifying desired properties and enforcing them through regularization terms in the loss function.  \\

Although the learned momenta clearly differ from the analytical momenta, we know from \cite{goldstein} that they must be related by a canonical transformation that preserves Hamiltonian structure. If the form of the analytical momenta is known, it can be written in terms of $\textbf{q}$ and $\dot{\textbf{q}}$ as $\textbf{p}_{a}=\textbf{p}_{a}(\textbf{q},\dot{\textbf{q}})$. Then, we can easily find that the canonical transformation connecting the learned momenta $\hat{\textbf{p}}$ and the analytical momenta $\textbf{p}_{a}$ takes the form:

\begin{equation}\label{ct}
    \textbf{p}_{a}=\textbf{p}_{a}(\textbf{q},\dot{\textbf{q}})=\textbf{p}_{a}(\textbf{q},\frac{\partial\hat{H}}{\partial \hat{\textbf{p}}})
\end{equation}

\noindent where the second equality comes from Eq. \ref{Heqs}. The second row of Figure \ref{momenta} compares the analytical momenta along the test trajectory with the learned momenta after being transformed as in Eq. \ref{ct}. Panels show the momenta of \textbf{e)} the mass-spring system, \textbf{f)} the double pendulum, \textbf{g)} spatial components of the rolling disk, and \textbf{h)} angular components of the rolling disk. As expected, by choosing the right canonical transformation, the analytical momenta are recovered to high precision, even if the untransformed learned momenta differ significantly.\\

To assess whether the learned momenta preserve the linearity in $\dot{\textbf{q}}$ of the analytical case, we analyze their behavior along a linear sweep in velocities. The sweep is performed by fixing $\textbf{q}$ and linearly progressing along the training set limits for $\dot{\textbf{q}}$, imposing Eq. \ref{c2} in the case of the rolling disk. We use $\textbf{q}=0$ for the mass spring system and the double pendulum, while $\textbf{q}=[0,0,0.4,0,0]$ is used for the rolling disk to ensure values fall inside training ranges. After doing a linear fit on the resulting learned momenta, the average coefficient of determination along all systems and all components is $R^{2}=0.9996$. This shows a clear linear behavior on $\dot{\textbf{q}}$ for all learned momenta, which is notable considering that no explicit linearity prior was used for the double pendulum or the rolling disk system.


\clearpage
\section{Conclusion}

We propose a general extension to Hamiltonian-based architectures that simultaneously learns a target system’s Hamiltonian structure, generalized momenta, and dynamics, when only data on the state and its derivatives are available. The proposed approach equips Hamiltonian-based architectures to deal with real world applications where obtaining data on the generalized momenta may be prohibitive. The novelty of our approach lies in the fact that it does not require any modifications to the loss function, does not rely on sequential data, and it is readily compatible with available approaches and additional priors. We also assess the performance of our architecture on systems with nonholonomic behaviors when no data on the momenta is available, which is unexplored in the literature. \\

A dissipative mass-spring system was first used, where the architecture was shown to not only accurately recover the dynamics of the system in the absence of data on the generalized momenta, but also to effectively reproduce the expected phase shift when encircling an exceptional point of the system. Testing on the double pendulum showed that the proposed
architecture is capable of producing accurate models of highly nonlinear systems with nontrivial  generalized momenta, when only data on the generalized coordinates and their derivatives are available. The rolling disk was used to assess compatibility of the proposed extension with more complex Hamiltonian architectures, achieving a highly accurate recovery of nonholonomic dynamics while allowing for seamless integration with the original architecture. \\

The learned momenta showed significant differences when compared to the analytical momenta, but behaved as expected when using the right canonical transformation. Notably, the learned momenta retained the linearity in $\dot{\textbf{q}}$ that is present in its analytical counterpart, even though no linearity prior was imposed for the double pendulum or rolling disk.\\

Possible directions for future work include designing and assessing priors that enforce desired behaviors on the generalized momenta, such as statistical independence; considering alternative approaches that also learn the generalized coordinates; and evaluating changes in performance when replacing neural networks by other modeling approaches.

\section*{Declaration of interest}
The authors have no conflicts to disclose.

\section*{Acknowledgments}
We acknowledge the support of the Natural Sciences and Engineering Research Council of Canada (NSERC), under their Discovery Grant Program (No. GR000989). 
This research was supported in part through computational resources and services provided by Advanced Research Computing at the University of British Columbia.

\section*{Declaration of generative AI and AI-assisted technologies in the manuscript preparation process}

During the preparation of this work, the authors used OpenAI's ChatGPT 5.5 Instant and 5.6 Luna, as well as Google's Gemini 3.5 Flash and 3.6 Flash to improve readability and language. After using these tools, the authors reviewed and edited the content as needed and take full responsibility for the content of the published article.

\bibliographystyle{elsarticle-num} 
\bibliography{cas-refs}

\clearpage
\appendix
\setcounter{table}{0}
\section{Training Parameters}
 
The training parameters used in Section \ref{TDS} are shown in Tables \ref{mshp}--\ref{diskhp}. Table \ref{diskranges} shows the uniform sampling intervals for the training set of the rolling disk.

\begin{table}[htbp]
\centering
\caption{Training parameters used for the mass-spring experiment.}
\label{mshp}
\begin{tabular}{
    p{0.26\textwidth}
    >{\centering\arraybackslash}p{0.135\textwidth}
    >{\centering\arraybackslash}p{0.135\textwidth}
    >{\centering\arraybackslash}p{0.135\textwidth}
}
\hline
\centering\textbf{Parameter} 
& \textbf{$\mathbf{p}$} 
& \textbf{$\mathbf{K}$ and $\mathbf{D}$}
& \textbf{$\mathbf{M}$} \\ \hline
Hidden layers       & 2                & 2                & N/A\\
Neurons per layer   & 30               & 10               & N/A\\
Activation function & Tanh             & Tanh             & N/A\\
Learning rate       & $1\times10^{-5}$ & $1\times10^{-4}$ & $1\times10^{-4}$\\
Epochs              & \multicolumn{3}{c}{5e4} \\
Training set        & \multicolumn{3}{c}{5e4 points} \\
Validation set      & \multicolumn{3}{c}{1e4 points} \\
Batches             & \multicolumn{3}{c}{100} \\ \hline
\end{tabular}
\end{table}

\begin{table}[htbp]
\centering
\caption{Parameters used for the double pendulum experiment.}
\label{dphp}
\begin{tabular}{lccc}
\hline
\multicolumn{1}{c}{\textbf{Parameter}} & \multicolumn{1}{c}{{\textbf{Value}}} \\ \hline
Hidden layers                            & 2 \\ 
Neurons per layer                        & 30 \\ 
Activation function                      & Tanh \\ 
Learning rate                            & 5e-5 \\ 
Epochs                                   & 3e6 \\
Training set                             & 4e3 points \\ 
Validation set                           & 1e3 points \\ 
Batches                                  & 1 \\ \hline
\end{tabular}
\end{table}

\begin{table}[htbp]
\centering
\caption{Parameters used for the rolling disk experiment.}
\label{diskhp}
\begin{tabular}{lcccc}
\hline
\multicolumn{1}{c}{\textbf{Parameter}} & \multicolumn{1}{c}{\textbf{$H$}} & \multicolumn{1}{c}{\textbf{$\bm{\lambda}$}} & \multicolumn{1}{c}{\textbf{$\textbf{A}$}} & \multicolumn{1}{c}{$\textbf{p}$} \\ \hline
Hidden layers                            & \multicolumn{1}{c}{2}                                    & \multicolumn{1}{c}{2}  & \multicolumn{1}{c}{2}                                         & 2                                    \\ 
Neurons per   layer                      & \multicolumn{1}{c}{400}                                  & \multicolumn{1}{c}{400}                                        & \multicolumn{1}{c}{100} & 400                                 \\ 
Activation   function                    & \multicolumn{1}{c}{Tanh}                                 & \multicolumn{1}{c}{Tanh} & \multicolumn{1}{c}{Tanh}                                      & Tanh                                 \\ 
Learning rate                            & \multicolumn{1}{c}{1e-5}                                 & \multicolumn{1}{c}{1e-5} & \multicolumn{1}{c}{1e-5}                                        & 1e-5                                 \\ 
Epochs                                   & \multicolumn{4}{c}{6e4}                                                                                                                                         \\ 
Training set                             & \multicolumn{4}{c}{4.9e4 points}                                                                                                                                  \\ 
Validation set                           & \multicolumn{4}{c}{1e3 points}                                                                                                                                    \\ 
Batches                                  & \multicolumn{4}{c}{200}                                                                                                                                            \\ \hline
\end{tabular}
\end{table}

\begin{table}[htbp]
\centering
\caption{Ranges of values for generated points for the rolling disk experiment.}
\label{diskranges}
\begin{tabular}{ccc}
\hline
\multicolumn{1}{l}{\textbf{Coordinate}} & \multicolumn{1}{l}{\textbf{Lower limit}} & \multicolumn{1}{l}{\textbf{Upper limit}} \\ \hline
$q_{1}$                                       & $-2.3$                                      & $2.3$                                       \\ 
$q_{2}$                                       & $-2.3$                                      & $2.3$                                       \\ 
$q_{3}$                                  & $0.3$                              & $0.5$                              \\ 
$q_{4}$                                    & $-0.3$                                         & $7$                                    \\ 
$q_{5}$                                    & $-0.3$                                         & $16$                                    \\ 
$\dot{q}_{1}$                                   & $-2.6$                                        & $2.6$                                         \\ 
$\dot{q}_{2}$                                   & $-2.6$                                        & $2.6$                                         \\ 
$\dot{q}_{3}$                              & $-0.2$                                      & $0.2$                                       \\ 
$\dot{q}_{4}$                                 & $1$                                    & $1.3$                                    \\ 
$\dot{q}_{5}$                                 & $2.4$                                     & $3$                                     \\ \hline
\end{tabular}
\end{table}

\end{document}